\documentclass[letterpaper, 10 pt, conference]{ieeeconf}

\IEEEoverridecommandlockouts
\usepackage[letterpaper, left=0.75in, right=0.75in, bottom=0.75in, top=0.75in]{geometry}

\usepackage{cite}
\usepackage{amsmath,amssymb,amsfonts,mathrsfs}
\usepackage{graphicx}
\usepackage{subcaption}
\usepackage{textcomp}
\usepackage{xcolor}
\usepackage{float}
\usepackage{setspace, layouts}

\usepackage{balance}
\usepackage{multirow}
\usepackage{multicol}
\usepackage{makecell}
\usepackage{array, caption, tabularx,  ragged2e,  booktabs}
\usepackage{CJKutf8}
\PassOptionsToPackage{hyphens}{url}
\usepackage{hyperref}  
\newcolumntype{P}[1]{>{\centering\arraybackslash}p{#1}}  
\newcolumntype{M}[1]{>{\centering\arraybackslash}m{#1}}  
\usepackage{fancyhdr}  

\usepackage{algorithm,algorithmic}

\usepackage{pifont}
\usepackage[T1]{fontenc}

\usepackage{threeparttable}
	
\def\BibTeX{{\rm B\kern-.05em{\sc i\kern-.025em b}\kern-.08em
    T\kern-.1667em\lower.7ex\hbox{E}\kern-.125emX}}

\begin{document}
\title{\LARGE \bf
Improved Robustness and Safety for Pre-Adaptation of Meta Reinforcement Learning with Prior Regularization
}

\author{Lu Wen$^{1}$,  Songan Zhang$^{2}$, H. Eric Tseng$^{2}$, Baljeet Singh$^{2}$, Dimitar Filev$^{2}$, Huei Peng$^{3}$
\thanks{$^{1}$Lu Wen is a Ph.D. Student in Mechanical Engineering, University
of Michigan, email: {\tt\small lulwen@umich.edu}}
\thanks{$^{2}$Songan Zhang, Eric Tseng, Baljeet Singh, Dimitar Filev are with Ford Research and Advanced Engineering.}
\thanks{$^{3}$Huei Peng is a professor of Mechanical Engineering, University of
Michigan, email: {\tt\small hpeng@umich.edu}}}

\maketitle
\thispagestyle{empty}
\pagestyle{empty}
\renewcommand{\footrulewidth}{0pt}
\renewcommand{\headrulewidth}{0pt}

\begin{abstract}
Meta Reinforcement Learning (Meta-RL) has seen substantial advancements recently. In particular, off-policy methods were developed to improve the data efficiency of Meta-RL techniques. \textit{Probabilistic embeddings for actor-critic RL} (PEARL) is a leading approach for multi-MDP adaptation problems. A major drawback of many existing Meta-RL methods, including PEARL, is that they do not explicitly consider the safety of the prior policy when it is exposed to a new task for the first time. Safety is essential for many real world applications, including field robots and Autonomous Vehicles (AVs). In this paper, we develop the PEARL PLUS (PEARL$^+$) algorithm, which optimizes the policy for both prior (pre-adaptation) safety and posterior (after-adaptation) performance. Building on top of PEARL, our proposed PEARL$^+$ algorithm introduces a prior regularization term in the reward function and a new Q-network for recovering the state-action value under prior context assumptions, to improve the robustness to task distribution shift and safety of the trained network exposed to a new task for the first time. The performance of PEARL$^+$ is validated by solving three safety-critical problems related to robots and AVs, including two MuJoCo benchmark problems. From the simulation experiments, we show that safety of the prior policy is significantly improved and more robust to task distribution shift compared to PEARL. 

\end{abstract}

\section{Introduction}
It is important for trained Artificial Intelligence (AI) to adapt to a new task relatively quickly with a small amount of data. Although Deep Reinforcement Learning (DRL) methods were shown to be powerful for complicated sequential decision making problems, they usually learn a separate policy for each specific task. In recent years, researchers tried to design AI that can solve multiple tasks, by either switching to the corresponding policy it already learned\cite{liaw2017composing}, or learning and adapting to the new task. The first approach requires that we can enumerate the tasks, which may not be possible or practical for some real-world applications. Moreover, learning to master each and every task often requires a lot of training data, making it data inefficient.

The second approach, which is to enable quick learning/adaptation as new tasks arise, is named Meta Reinforcement Learning (Meta-RL)\cite{wang2016learning}. Meta-RL techniques aim to gain ``general learning" for better dealing with new tasks by leveraging learned experience from previous related tasks. This requires that these tasks share some common features. For instance, in autonomous driving, merging into a highway from a ramp and entering a round-about both involve the Autonomous Vehicle (AV) agent choosing a gap, adjusting its velocity, and making the move, while the exact road geometry and behaviors of the adjacent vehicles may vary.   

How to leverage learning for a family of tasks sharing similar structures, instead of simply doing well in a single task is still an open topic. In \cite{finn2017model,antoniou2019train,nichol2018first}, the agent is trained to find the parameters $\theta$ of its policy, so that when the agent takes a few gradient steps on that $\theta$, it will converge toward $\theta^*$ which is optimal for a new task. In this case, the common structure of different tasks is implicitly represented by the initial parameter $\theta$. 
Another approach is to learn latent contexts among tasks and the context acts as part of the input to the policy network. For example, in \cite{rakelly2019efficient} Rakelly et al. developed a data efficient off-policy Meta-RL method called Probabilistic Embeddings for  Actor-critic  meta-RL (PEARL), performing online probabilistic filtering of the latent task variables to infer how to solve a new task from a small amount of new data. While the off-policy nature of PEARL enable data efficiency when adapting to a new task, it does require on-line collecting the context and corresponding latent variables of a new task with prior policy (the policy conditioned on the prior distribution). As a result, the performance is not guaranteed when encountering a new task for the first time. This poses a serious challenge for applications like autonomous driving where the agent cannot afford to perform very poorly. In other words, it is important for the prior (pre-adaptation) policy to maintain robustness with respect to task distribution shift.

Safety in RL is not a new problem and a lot of research has been done \cite{garcia2015comprehensive, wen2021safe}, but the safety of Meta-RL before adaptation has not received enough attention. In this paper, we consider a policy as safe if it would not lead to a priorly-defined unsafe state. Robustness in meta-learning has also been studied in multiple recent works. The work in \cite{zugner2019adversarial} explores robustness to perturbations in the task samples, \cite{rafique2021weakly} deals with imbalances in the number of samples per task instance, and \cite{collins2020task} proposes a solution to improve the robustness to shifts in the task distribution between meta-training and meta-testing. In this paper our discussion is on 'pre-adaptive task-robustness' in the sense that it performs well before adaptation despite the task-distributional shift between prior assumption and the real. All of our experimental studies are safety-critical cases and we will consider a prior policy pre-adaptive task-robust if its ‘safety violation’ rate does not exceeds an absolute rate of 0.1\%. We develop PEARL+, a Meta-RL method based on PEARL\cite{rakelly2019efficient} to 
improve safety and robustness for pre-adaptation of Meta-RL. We introduce a prior regularization term and a new Q-network, which doesn't back-propagate into the context encoder, to explicitly consider the robustness and safety of the prior policy  without sacrificing after-adaptation performance.

The rest of the paper is organized as follows. We first describe the related work in Section \ref{RelatedWork}. Then the preliminaries of Meta-RL and the PEARL method will be introduced in Section \ref{Preliminaries}. In Section \ref{PEARL$^+$}, we present the details of our new method PEARL$^+$. We study the performance of PEARL$^+$ on three applications and the results are shown in Section \ref{ExperimentsAndResults}. Finally, the paper is concluded in Section \ref{Conclusion}.

\section{Related Work}
\label{RelatedWork}
The current state-of-the-art Meta-RL algorithms can be categorized into four types. 

\textit{RNN-based Meta-RL.} 
In these methods (e.g., \cite{wang2016learning}\cite{duan2016rl}), a Recurrent Neural Network (RNN) structure is used, the experience from previous tasks are preserved in the RNN parameters.  The agent starts with no memory/knowledge when it starts to adapt to a new task. However, RNN based policy has no mathematical convergence proof and is believed to be vulnerable to meta-overfitting \cite{metalearning_tutorial}.

\textit{Gradient-based Meta-RL.}
This type of Meta-RLs learn a policy initialization that can quickly adapt to a new task after few policy gradient steps. They usually separate learning into two loops: inner loop for policy adaptation, and outer loop for policy updates\cite{zhang2021quick}. In most existing work, optimizations are through gradient descents for both loops\cite{finn2017model}\cite{nichol2018first}. One notable exception is \cite{mendonca2019guided}, which implements behavior cloning in the inner loop to reduce the need for task exploration. Due to their on-policy nature, these methods usually need a lot of data and parameter-tuning\cite{metalearning_tutorial}.

\textit{Model-based Meta-RL}
These methods learn a predictive model, represented by a deep neural network, and find model parameters that are sensitive to variations among the tasks\cite{belkhale2021model}\cite{lin2020model}. The action to take is then determined through optimization methods such as model-predictive control (MPC). Due to their model-based nature, these methods require less data and can adapt more quickly. However, they require relatively high computation cost for the online optimization process to get actions\cite{metalearning_tutorial}.

\textit{Context-based Meta-RL}
In this category, the meta-agents learn the latent variables of a distribution of tasks, which the policy uses to adapt to new tasks \cite{rakelly2019efficient,gupta2018meta, fakoor2019meta, kirsch2019improving, zhou2020online}. Our method falls into this category. It is worth noting that the approach in this category allows off-policy learning. That is, the RL learning does not require the latest updated policy to interact with the environment. Rather, it can leverage experiences from other policies, such as learning from replaying buffer samples from old policy as used in \cite{fakoor2019meta}. Examples of this category include \cite{kirsch2019improving}\cite{zhou2020online} where their off-policy Meta-RL algorithms were developed by decoupling the task inference from the policy training.

Regardless of the categories, most of the meta RL literature has focused on data efficiency and fast adaptability. However, in some applications, the performance of the prior policy (i.e. prior to adaptation) may be critical to avoid failures as these cases cannot tolerate mischief.
For example, for automated vehicles, the factory installed default setting can be considered as the prior policy, while vehicles may be sold and driven in different cities/countries with a different driving culture/roadmanship, which can be considered as different tasks (MDPs).  In this case, the safety performance of the prior policy (factory setting) cannot be overlooked. We were not aware of any Meta-RL methods that explicitly considered the safety of the prior policy at the beginning of the meta testing, or when it is first exposed to a new task in the real-world.  This lack of assurance of robust initial performance is a concern of Meta-RL application to safety-critical tasks.

\section{Preliminaries}
\label{Preliminaries}
In this section, we will briefly review the formulation of our meta reinforcement learning problem, and introduce the PEARL algorithm which we build upon and enhance.
\subsection{Meta Reinforcement Learning}
Following the traditional Meta-RL problem formulations \cite{finn2017model}, we consider a distribution of tasks $p(\mathcal{T})$ that we want our model to adapt to, where each task $\mathcal{T}_i$ is modeled as a Markov decision process (MDP), consisting of a finite set of states $S$, actions $\mathcal{A}$, a transition function $P_i$, and a bounded reward function $R_i$, and the discount factor $\gamma_i$, i.e., $\mathcal{T}_i=(S, A, P_i, R_i, \gamma_i)$, where the state space and action space are same across tasks.

In Meta-RL, the goal is to train a meta-learner $\mathcal{M}(*)$ which learn from a large number of tasks (known as meta-training tasks) from a given distribution, with the hope that it can generalize to new tasks or new environments in the same distribution.  In the meta-testing process, the meta-learner's ability to adapt to new tasks (known as meta-testing tasks) sampled from the same distribution will be verified. In a mathematical form, the meta-learner $\mathcal{M}$ with parameters $\theta$ can be optimized by:
\begin{equation}
\label{eq:1}
    \mathcal{M}_{\theta} := \arg\max_{\theta}\sum_{i=1}^{N}\mathcal{L}_{\mathcal{T}_i}(\phi_i),
\end{equation}
\begin{equation}
\label{eq:2}
    \phi_i = f_{\theta}(\mathcal{T}_i).
\end{equation}
where $f$ is the adaptation operation, and $\mathcal{L}$ denotes the objective function, for which we use the expected reward function:
\begin{equation}
\label{eq:Meta-RL}
    \mathcal{L}(\theta) = \mathbf{E}_{\pi_{\theta}}\left[\sum_{t=0}^{\infty}\gamma_i^t r_i(t)\right]
\end{equation}

Facing a new task $\mathcal{T}_i$, the meta-learner adapts from parameter $\theta$ to $\phi_i$ after the adaptation operation in Eq. \ref{eq:2}. The adaptation function $f_{\theta}$ varies between methods. For example, in the MAML algorithm\cite{finn2017model} it consists of gradient descent, while in PEARL it consists of encoding experience to, and inference on, posterior task embeddings. The meta-learner optimizes its parameters to maximize the sum of objective values collected from performing on meta-training tasks with corresponding adapted parameter $\phi$ as Eq. \ref{eq:1}.

\subsection{PEARL algorithm}
The PEARL\cite{rakelly2019efficient} algorithm is an off-policy Meta-RL algorithm that disentangles task inference and control. PEARL captures knowledge about how the current task should be performed in a latent probabilistic context variable $Z$, for which it conditions the policy as $\pi_{\theta}(\mathbf{a|s,z})$ to adapt to the task. In the meta-training process, PEARL leverages data from a variety of training tasks to train a probabilistic encoder $q(\mathbf{z}|\mathbf{c})$ to estimate the posterior $p(\mathbf{z}|\mathbf{c})$, where $\mathbf{c}$ refers to \textit{context}, and $\mathbf{c}_{i}=(\mathbf{s}_i, \mathbf{a}_i, r_i, \mathbf{s}_i')$ denotes one transition tuple. 

The objective optimized by PEARL is:
\begin{equation}
\label{eq:pearl-obj}
    \mathbb{E}_{\mathcal{T}} [\mathbb{E}_{\mathbf{z}\sim q(\mathbf{z}|\mathbf{c}^{\mathcal{T}})} [R(\mathcal{T},\mathbf{z})+\beta D_{KL}(q(\mathbf{z}|\mathbf{c}^{\mathcal{T}})||p_0(\mathbf{z}))] ],
\end{equation}
where $p_0(\mathbf{z})$ is a unit Gaussian prior over the context variable $Z$. The KL divergence term can be interpreted as the result of a variational approximation to an information bottleneck\cite{alemi2016deep} that constrains $\mathbf{z}$ to contain only necessary information to adapt to the task, mitigating the overfitting issue.

PEARL is built on top of the soft actor-critic (SAC) algorithm\cite{haarnoja2018soft}, which exhibits good sample efficiency and stability. The inference network (also known as the encoder) $q_{\phi}(\mathbf{z|c})$ is optimized jointly with the actor $\pi_{\theta}(\mathbf{a|s,z})$ and critic $Q_{\theta}(\mathbf{s,a,z})$. If we use the state-action value as the metric to train the encoder, the critic and actor loss can be written respectively as Eq. \ref{eq:pearl-critic} and Eq. \ref{eq:pearl-actor}:
\begin{equation}
\label{eq:pearl-critic}
    \mathcal{L}_{critic} = \mathbf{E}_{\substack{(\mathbf{s,a,},r,\mathbf{s'})\sim\mathcal{B} \\                                                          \mathbf{z}\sim q_{\phi}(\mathbf{z|c})}}
                        [Q_{\theta}(\mathbf{s,a,z})-(r+\bar{V}(\mathbf{s',\bar{z}}))]^2.
\end{equation}
where $\bar{V}$ is a target value network and $\mathbf{\bar{z}}$ means that gradients are not back-propagated through the network.

\begin{equation}
\label{eq:pearl-actor}
    \mathcal{L}_{actor} = 
 \mathbf{E}_{\substack{\mathbf{s}\sim\mathcal{B}, \mathbf{a}\sim\pi_{\theta} \\                                                          \mathbf{z}\sim q_{\phi}(\mathbf{z|c})}}
                        \left[D_{KL}\left(\pi_{\theta}(\mathbf{a|s,\bar{z}})\left|\right|\frac{\exp{(Q_{\theta}(\mathbf{s,a,\bar{z}}))}}{\mathcal{Z}_{\theta}(\mathbf{s})}\right)\right].
\end{equation}
where the partition function $\mathcal{Z}_{\theta}$ normalizes the distribution, and it does not contribute to the gradient with respect to the policy.

The data used to infer $q_{\phi}(\mathbf{z|c})$ is distinct from that used to construct the critic loss. As illustrated in Fig. \ref{fig:diagram}, the context data sampler $\mathcal{S}_\mathbf{c}$ samples uniformly from the most recently collected data batch, while the actor and critic are trained with data drawn uniformly from the entire replay buffer.

During meta-testing, PEARL samples the context variables from the unit Gaussian prior and hold constant in an episode to do temporally-extended exploration when faced with an unseen task. Later, the latent context variable of the new task is inferred from the gathered experience with the inference network $q_{\phi}(\mathbf{z|c})$.

\section{PEARL$^+$ and the prior regularization term}
\label{PEARL$^+$} 
Most of the Meta-RL algorithms proposed in the literature focused on optimizing the performance adapting to specific tasks, but did not explicitly consider the performance before adaptation. For some applications like robots and automated vehicles, any fatal mistakes are unacceptable, including before adaptation. In this section, we'll explain details of our algorithm PEARL$^+$, which is a direct enhancement to PEARL.

In PEARL's configuration, the prior over context variable $z$ is denoted as $z_0$, which is assumed to follow an identical independent distribution. However, this assumption does not hold for real tasks and environments. Since the previous formulation (Eq. \ref{eq:pearl-obj}) only optimizes on the posterior return, the policy trained in this way doesn't perform well under on the prior assumption before adaptation with regard to the distribution shift. We propose to solve the problem by introducing a prior regularization term, to calculate which we create a new Q-Network for state-action value estimation under the prior context assumption. 

The prior regularization is formulated as the expected return depending on the random context variable following the prior unit Gaussian distribution $p(\mathbf{z_0})$. The objective function of PEARL$^+$ can then be formulated as:
\begin{multline}
\label{eq:pearl+obj}
     \mathbb{E}_{\mathcal{T}} [\mathbb{E}_{\mathbf{z}\sim q(\mathbf{z}|\mathbf{c}^{\mathcal{T}})} [R(\mathcal{T},\mathbf{z})+\beta D_{KL}(q(\mathbf{z}|\mathbf{c}^{\mathcal{T}})||p_0(\mathbf{z}))] \\
     + \alpha \mathbb{E}_{\mathbf{z}\sim p(\mathbf{z}_0)} [R(\mathcal{T},\mathbf{z})]]  
\end{multline}
where $\alpha>0$ is the trade-off coefficient.  

The policy is trained to maximize the return which considers both prior (before adaptation) and posterior (after adaptation) of the context variable. Similar to the exploration-exploitation trade-off: emphasizing prior regularization performance will decrease performance of the posterior context. Note that the prior regularization term does not depend on the inference network.  Therefore, it will not affect the training of the context encoder $q_{\phi}$.  However, we can train the policy to be more robust by considering the mismatch with the real context.
\begin{figure}[h]
	\centering
	\includegraphics[width=0.34\textwidth]{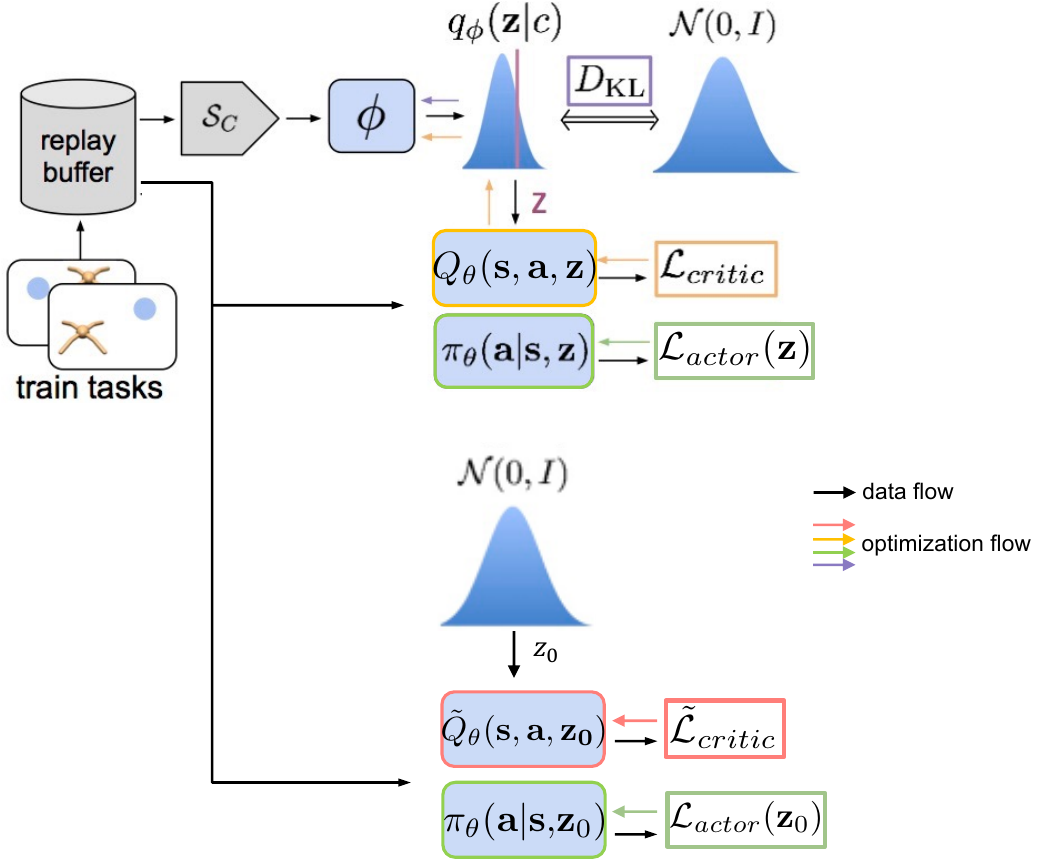}
	\caption{The PEARL$^+$ diagram}
	\label{fig:diagram}
\end{figure}

Based on the structure of PEARL, we introduce a new Q-network $\Tilde{Q}_{\theta}$ to recover the action-state value with a prior context assumption $\mathbf{z_0}$. As shown in Fig. \ref{fig:diagram}, we update the $\Tilde{Q}_\theta$ network by gradients computed from loss $\Tilde{\mathcal{L}}_{critic}$ (red line) with data sampled from the whole replay buffer. Similar to PEARL, the policy of PEARL$^+$ $\pi_\theta$ is also built on top of the Soft Actor Critic (SAC) algorithm, but now has two optimization sources (both shown in green lines): the original $\mathcal{L}_{actor}(\mathbf{z})$ computed with the posterior context, and an additional $\mathcal{L}_{actor}(\mathbf{z_0})$ computed with 
the prior context. They share the same set of data sampled from the whole replay buffer. The encoder is optimized by $\mathcal{L}_{critic}$ (yellow line) and $D_{KL}$ (purple line), which is the same as PEARL. 

We summarize the meta-training procedure of PEARL$^+$ in Algorithm \ref{alg:pearl+}. The meta-testing procedure is the same as PEARL, and therefore will not be described in this paper. 
\begin{algorithm}[ht]
 \caption{PEARL$^+$ Meta-training}
 \begin{algorithmic}[1]
 \label{alg:pearl+}
 \renewcommand{\algorithmicrequire}{\textbf{Require:}}
 \REQUIRE Batch of training tasks $\{\mathcal{T}_i\}_{i=1,\dots T}$ from $p(\mathcal{T})$, learning rates $\alpha_1, \alpha_2, \alpha_3$

\STATE Initialize replay buffers $\mathcal{B}^i$ for each training task
\WHILE{not done}
    \FOR {each $\mathcal{T}_i$}
        \STATE Initialize context $\mathbf{c}^i=\{\}$
        \FOR {$k=1,\dots,K$}
            \STATE Sample $\mathbf{z}\sim q_{\phi}(\mathbf{z}|\mathbf{c}^i)$
            \STATE Gather data from $\pi_{\theta}(\mathbf{a}|\mathbf{s,z})$ and add to $\mathcal{B}^i$
            \STATE Update $\mathbf{c}^i=\{(\mathbf{s}_j, \mathbf{a}_j,\mathbf{s'}_j,r_j)\}_{j:1\dots N}\sim\mathcal{B}^i$
        \ENDFOR
    \ENDFOR
    \FOR {step in training steps}
        \FOR {each $\mathcal{T}_i$}
            \STATE Sample context $\mathbf{c}^i\sim \mathcal{S}_C(\mathcal{B}^i)$ and RL batch $b^i\sim\mathcal{B}^i$
            \STATE Sample $\mathbf{z}\sim q_{\phi}(\mathbf{z}|\mathbf{c}^i)$
            \STATE Sample $\mathbf{z}_0\sim N(0,1)$
            \STATE $\mathcal{L}_{actor}^i=\mathcal{L}_{actor}(b^i, \mathbf{z})+ \alpha \mathcal{L}_{actor}(b^i,\mathbf{z}_0)$
            \STATE $\mathcal{L}_{critic}^i=\mathcal{L}_{critic}(b^i, \mathbf{z})$
            \STATE $\Tilde{\mathcal{L}}_{critic}^i=\Tilde{\mathcal{L}}_{critic}(b^i, \mathbf{z}_0)$
            \STATE $\mathcal{L}_{KL}^i=\beta D_{KL}(q(\mathbf{z}|\mathbf{c}^i)||r(\mathbf{z}))$
        \ENDFOR
        \STATE $\phi\xleftarrow{}\phi-\alpha_1\nabla{\phi}\sum_i(\mathcal{L}_{critic}^i + \mathcal{L}_{KL}^i)$
        \STATE $\theta_{\pi}\xleftarrow{}\theta_{\pi}-\alpha_2\nabla{\theta}\sum_i \mathcal{L}_{actor}^i$
        \STATE $\theta_{Q}\xleftarrow{}\theta_{Q}-\alpha_3\nabla{\theta}\sum_i \mathcal{L}_{critic}^i$        
        \STATE $\theta_{\Tilde{Q}}\xleftarrow{}\theta_{\Tilde{Q}}-\alpha_3\nabla{\theta}\sum_i \Tilde{\mathcal{L}}_{critic}^i$         
        
    \ENDFOR
 \ENDWHILE
 \end{algorithmic} 
 \end{algorithm}

\section{Experiments and results}
\label{ExperimentsAndResults}
In this section, we introduce the three experiments used to examine the performance of the PEARL$^+$ algorithm. All three examples are in the field of robotics and autonomous vehicles, and their prior safety performance (after training, but before adaptation to a new task) is important.
\subsection{Experimental setup}
The three experiments are: two continuous control tasks with robots in the MuJoCo simulator\cite{todorov2012mujoco}, and one discrete decision-making problem in highway ramp merging.
    \subsubsection{MuJoCo}{Walker-2D-Vel, Ant-3D-Vel.}
    \begin{figure}[h]
        	\centering
        	\includegraphics[width=0.25\textwidth]{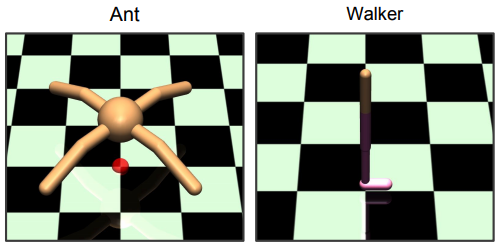}
        	\label{fig:mujoco_tasks}
        	\caption{MuJoCo robots: ant(3D), and walker(2D)}
        \end{figure}
    We choose two MuJoCo examples from \cite{finn2017model} and \cite{rakelly2019efficient} because the safety metrics of both are well defined, namely health status. The health status is evaluated by how reasonable the moving pose of the robot is on the bionics. Specifically, healthy pose means moving like a human's leg without jumping, falling or leaning for the MuJoCo Walker, and acting like a four legged insect with no jumping or slipping for the MuJoCo Ant. The episode terminates when the pose is unhealthy or reaches the maximum episodic length. These two locomotion tasks require adaptation to improve the reward functions: walking velocity for the Walker-2D-Vel, and moving direction and velocity for the Ant-3D-Vel. For details of the environment configurations, please refer to \cite{rakelly2019efficient}.
        
    \subsubsection{Self-driving}{Highway Merge}
    \begin{figure}[H]
        	\centering
        	\includegraphics[width=0.33\textwidth]{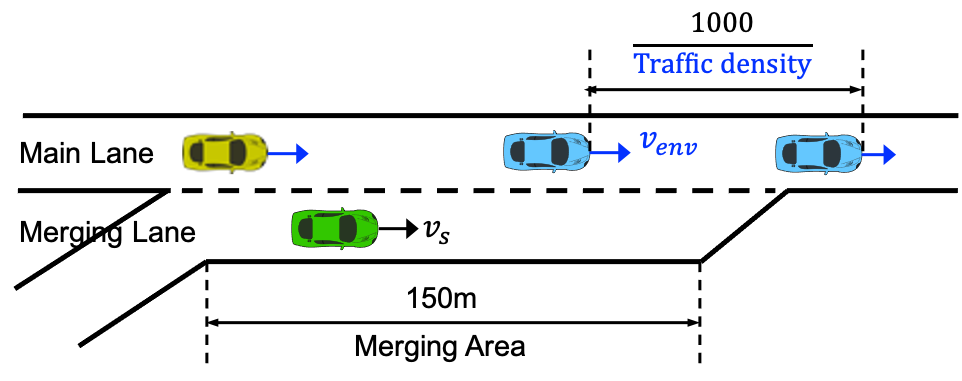}
        	\caption{Highway-Merging Scenario}
        	\label{fig:Merge-env}
        \end{figure}
        
    Meta-RL has been applied to solve decision-making problems in autonomous driving \cite{zhang2021quick}\cite{wang2021meta}\cite{jaafra2019context}, but none of the references we had reviewed studied the prior safety, which is an important issue for AVs.
    
    As shown in Fig. \ref{fig:Merge-env}, we consider a scenario for an automated vehicle: the ego vehicle (the green vehicle in Fig. \ref{fig:Merge-env}) is expected to take a mandatory lane change to merge into the main lane before the merging area ends. The vehicles driving in the main lane under normal situations would follow the Intelligent Driver Model (IDM)\cite{treiber2000congested} for their longitudinal dynamics. Since there is only one lane on the highway, lateral dynamics are not considered.  When the ego vehicle is still on the ramp and hasn't merged in, the vehicle on the main lane right behind it (the yellow vehicle in Fig. \ref{fig:Merge-env}) will switch from the IDM car-following model to the Hidas' model\cite{HIDAS200537} interacting with the merging ego vehicle. 
    
    \begin{figure}[h]
    	\centering
    	\includegraphics[width=0.33\textwidth]{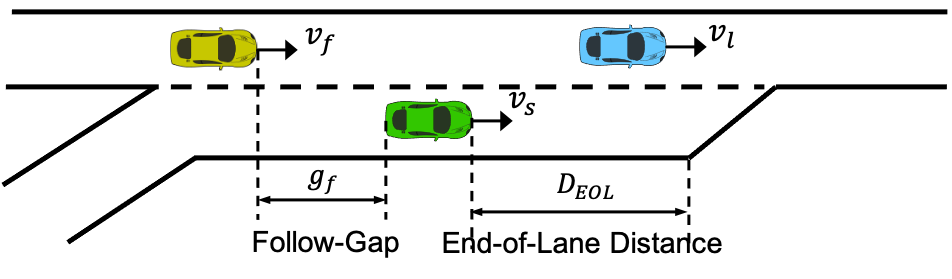}
    	\caption{Hidas's merging model. The green, yellow, and blue vehicle are the subject, follower, and lead vehicles, respectively, with velocity $v_s$, $v_f$ and $v_l$.}
    	\label{fig:hidas}
    \end{figure}
    In the Hidas model, the main lane vehicle behind the merging ego vehicle assesses if the “follow gap” (shown in Fig. \ref{fig:hidas}) would be sufficient/feasible for merging (i.e. larger than a minimum gap) should it yield and slow down willingly. The willingness is defined by maximum braking $b_f$ and range of speed decrease $Dv$. If not, the main lane vehicle will ignore the merging vehicle. The feasibility of the follower to slow down is calculated as follows:
    \begin{equation}
    \label{eq:min_b}
        Dt = Dv/b_f
    \end{equation}
    \begin{equation}
    \label{eq:min_b_1}
        g_{f,sld} = g_f-(v_fDt-b_fDt^2/2)+v_sDt
    \end{equation}
        \begin{equation}
    \label{eq:min_b_1}
        g_{f,min} = g_{min}+
        \begin{cases}
             c(v_f-v_s),&\text{if } v_f>v_s  \\
             0,& \text{otherwise} 
        \end{cases}
    \end{equation}
    The follower vehicle slows down if $g_{f,sld}>g_{f,min}$, or else it follows its lead vehicle on the main lane with the IDM model instead.

    The Hidas' interactive model parameters used in our experiment were calibrated from the collected video data by\cite{HIDAS200537} and are as follows: 
    \begin{itemize}
        \item Maximum speed decrease, $Dv=2.7 \text{m/s}(\sim 10 \text{km/h})$.
        \item Minimum safe constant gap $g_\text{min}=2.0 \text{m}$.
        \item Acceptable gap parameter $c=0.9$.
        \item Acceptable deceleration rate $b_f=1.5 \text{m/s$^2$}$
    \end{itemize}
    
    This self-driving task requires the Meta-RL to adapt to the transition probability functions. The MDP varieties considered here are marked in blue in Fig. \ref{fig:Merge-env}, and are sampled within their ranges to represent different driving environments:
    \begin{itemize}
        \item traffic density $p \in [30, 50]$ veh/km; 
        \item traffic speed $v_{env} \in [50, 70]$ mph;
    \end{itemize}
    
    The state-space $S\subseteq R^n$ of the learning agent includes the ego-vehicle-related states: [$D_{EOL}$, $y$, $v_x$,$v_y$], and surrounding-vehicle-related states: [$f_e^i$, $\Delta x^i$, $\Delta y^i$, $\Delta v_x^i$, $\Delta v_y^i$ ]. We consider 4 nearest surrounding vehicles on the main lane in the overall state space $S\subseteq R^{24}$. The five discrete actions of the learning agent include: accelerate ($1.5m/s^2$), decelerate ($-1.5m/s^2$), cruise (IDM), left lane-change, and right lane-change.
    
    The reward function is designed to consider multiple factors: relative velocity, distance to the front vehicle, merging safety (evaluated by the deceleration of the rear vehicle), penalty for collisions, and action cost:
    \begin{equation}
    \begin{split}
    R(s,a)= & \quad \alpha\left|v_x-v_{ego,target}\right|\\
            & + \beta \left|\Delta x^{front}-\frac{1000}{\text{traffic density}}\right|\\
            & + \gamma a_{rear,merge} + r_{collision} + r_{a},
    \end{split}
    \end{equation}
    where $\alpha$, $\beta$, $\gamma$ are the weighting coefficients. $a_{rear,merge}$ is the deceleration of the rear vehicle and is only applied at the time-step when the ego vehicle merges into the main lane.

\subsection{Safety and performance comparison}
\begin{figure*}[h]
	\centering
    \begin{subfigure}[b]{.49\textwidth}
        \includegraphics[width=\textwidth]{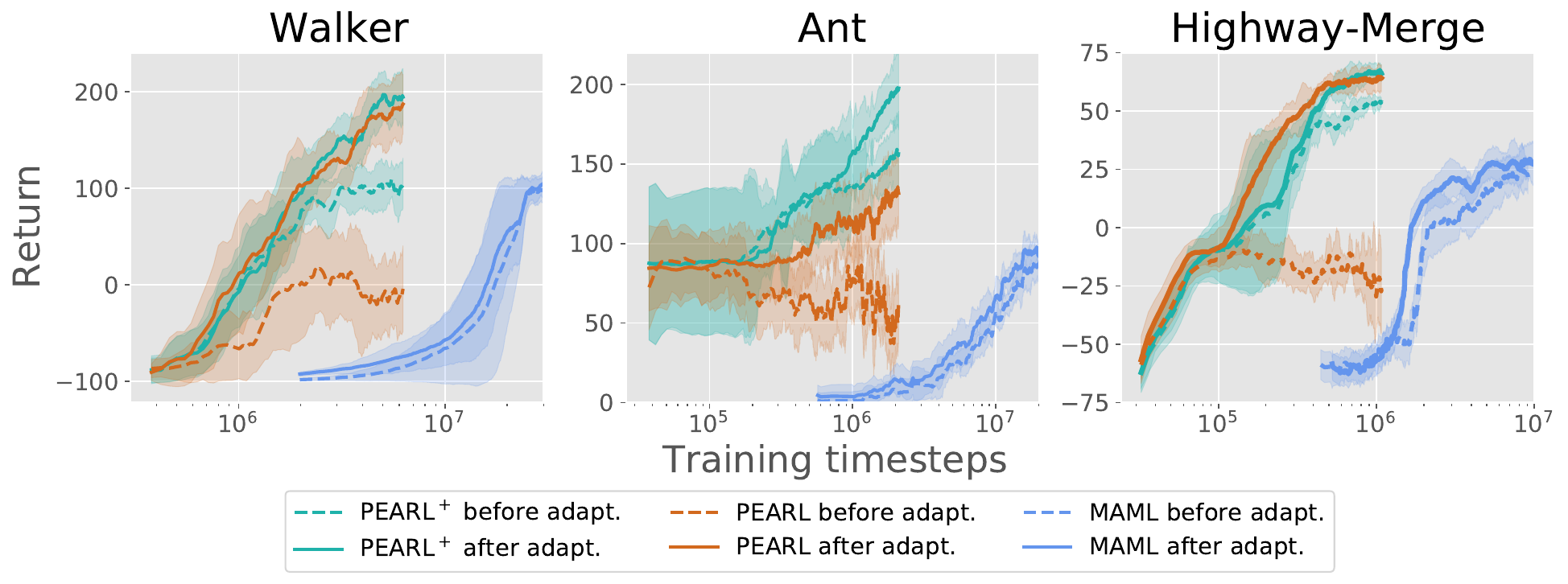}
        \subcaption{Meta Training Results}
        \label{fig:meta-train-results}
    \end{subfigure}
    \begin{subfigure}[b]{.49\textwidth}
        \includegraphics[width=\textwidth]{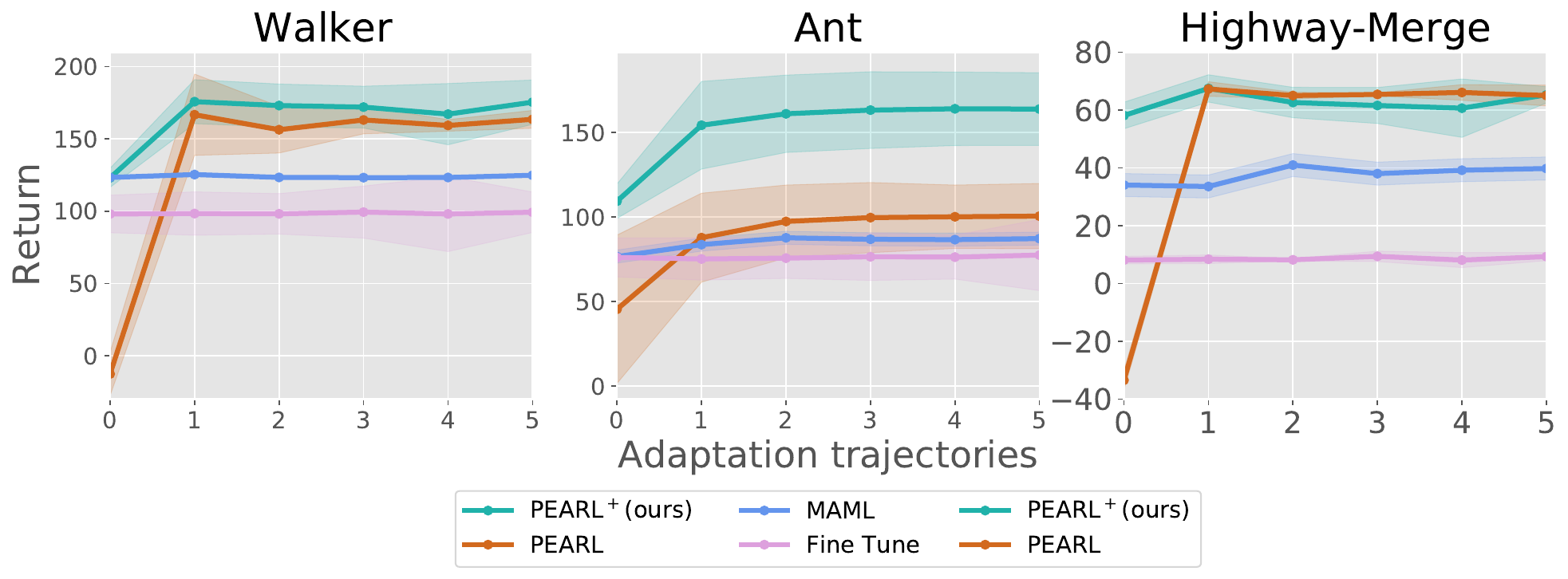}
        \subcaption{Meta Testing Results}
        \label{fig:meta-test-results}
    \end{subfigure}
	\caption{Meta Training Results}
\end{figure*}

The average episodic return on meta-training tasks throughout the training process is shown in Fig. \ref{fig:meta-train-results}. Upon the completion of meta-training, we further compare the performance among different algorithms along the adaptation trajectories (See Fig. \ref{fig:meta-test-results}).  The unhealthy/crash rates corresponding to Fig \ref{fig:meta-test-results} are shown in Table \ref{table:walker2d},\ref{table:ant3d},
\ref{table:crash_rate_merge}. The results are discussed below.

\textbf{Prior safety and performance.}
From Fig. \ref{fig:meta-test-results} and Table \ref{table:walker2d},\ref{table:ant3d},
\ref{table:crash_rate_merge} we can see that, even though PEARL and PEARL${^+}$ both show increasing reward with adaptation, PEARL${^+}$ has much higher reward and significantly lower unhealthy/crash rate before adaptation (using their policies conditioned on the prior inference of the task). As unhealthy poses or crashes carry very heavy penalty in all three experiments, substantially better performance implies improved safety. 


\textbf{Posterior adaptation and performance.}
Fig. \ref{fig:meta-test-results} and Fig. \ref{fig:meta-train-results} show that in all experiments, the episodic reward of PEARL${^+}$ after sufficient adaptation is very close to or better than that of PEARL. In other words, PEARL+ has a better prior performance, and this is not achieved without sacrificing posterior performance compared with PEARL. 


\textbf{Training efficiency.} From Fig. \ref{fig:meta-train-results} we can see that PEARL$^+$ is similar to PEARL in terms of sample efficiency and time efficiency, since they're both built upon the same off-policy algorithm SAC. PEARL$^+$ outperforms MAML (an on-policy algorithm) in terms of sample efficiency by a factor of 10-100X in our experiments.

To draw a conclusion, PEARL${^+}$ improves the policy's performance based on prior, without sacrificing its adaptation performance. In the robotics and autonomous driving tasks described above, PEARL${^+}$ learns a meta policy that is safe before adaptation (the main improvement) and achieves similar or better performance during and after adaptation compared with the original PEARL algorithm.

\begin{table}[ht]
\caption{Meta-Testing Unhealthy/Crash Rate Results}
\begin{subtable}{.5\textwidth}
    \centering
	\begin{tabular}[h]{m{0.1\linewidth}|M{0.14\linewidth}|M{0.14\linewidth}|M{0.14\linewidth}|M{0.16\linewidth}}
		\hline
		            & Fine Tune             & MAML                  &   PEARL   & PEARL${^+}$ (ours)\\ 
		\hline
		0 traj.     &$\sim\textbf{0.0\%}$  &$2.00\%$                &$60.7\%$   &$\sim\textbf{0.0\%}$\\ 
        \hline
		1 traj.     &$\sim\textbf{0.0\%}$  &$1.67\%$                &$2.9\%$    &$\sim\textbf{0.0\%}$\\
        \hline
		3 traj.     &$1.11\%$               &$2.67\%$                &$0.7\%$    &$\sim\textbf{0.0\%}$\\
        \hline
		5 traj.     &$1.67\%$               &$1.33\%$  & $\sim\textbf{0.0\%}$  & $\sim\textbf{0.0\%}$\\
		\hline
	\end{tabular}
	\subcaption{Walker2D-vel: Unhealthy Rates}
	\label{table:walker2d}
\end{subtable}
\begin{subtable}{.5\textwidth}
    \centering
	\begin{tabular}[h]{m{0.1\linewidth}|M{0.14\linewidth}|M{0.1\linewidth}|M{0.16\linewidth}|M{0.18\linewidth}}
		\hline
		          & Fine Tune & MAML      &   PEARL   & PEARL${^+}$ (ours)\\ 
		\hline
		0 traj.  &  $13.3\%$  &  $3.33\%$ &$33.3\%$ &$\sim\textbf{0.0\%}$\\ 
        \hline
		1 traj.  &  $13.1\%$  & $6.67\%$ &$\sim\textbf{0.0\%}$        &$\sim\textbf{0.0\%}$\\
        \hline
		3 traj. &   $12.7\%$  &  $6.67\%$ &$\sim\textbf{0.0\%}$        &$\sim\textbf{0.0\%}$\\
        \hline
		5 traj.  &  $13.9\%$  & $3.33\%$  &$\sim\textbf{0.0\%}$   &$\sim\textbf{0.0\%}$ \\
		\hline
	\end{tabular}
	\caption{Ant3D-vel: Unhealthy Rates}
	\label{table:ant3d}
\end{subtable}
\begin{subtable}{.5\textwidth}
    \centering
	\begin{tabular}[h]{m{0.12\linewidth}|M{0.1\linewidth}|M{0.1\linewidth}|M{0.16\linewidth}|M{0.18\linewidth}}
		\hline
		          & Fine Tune & MAML      &   PEARL   & PEARL${^+}$ (ours)\\ 
		\hline
		0 traj.  &  $14.5\%$  &  $19.0\%$ & $27.48\%$  & $\sim\textbf{0.04\%}$\\ 
        \hline
		1 traj.  &  $14.8\%$  & $15.3\%$ & $0.86\%$     & $\sim\textbf{0.00\%}$\\
        \hline
		3 traj. &   $14.1\%$  &  $18.3\%$ &  $0.12\%$    & $\sim\textbf{0.00\%}$\\
        \hline
		5 traj.  &  $13.7\%$  & $19.7\%$  & $\sim\textbf{0.00\%}$  & $\sim\textbf{0.00\%}$\\
		\hline
	\end{tabular}
	\caption{Highway-Merge: Crash Rates}
    \label{table:crash_rate_merge} 
\end{subtable}
\end{table}

\subsection{Visualization of prior policy exploration}
In this section, we present example simulation results to visualize how PEARL$^+$ benefits the decision-making process compared with PEARL, with focus on the Highway-Merge task. 

\begin{figure}[ht]
	\centering
	\includegraphics[width=0.43\textwidth]{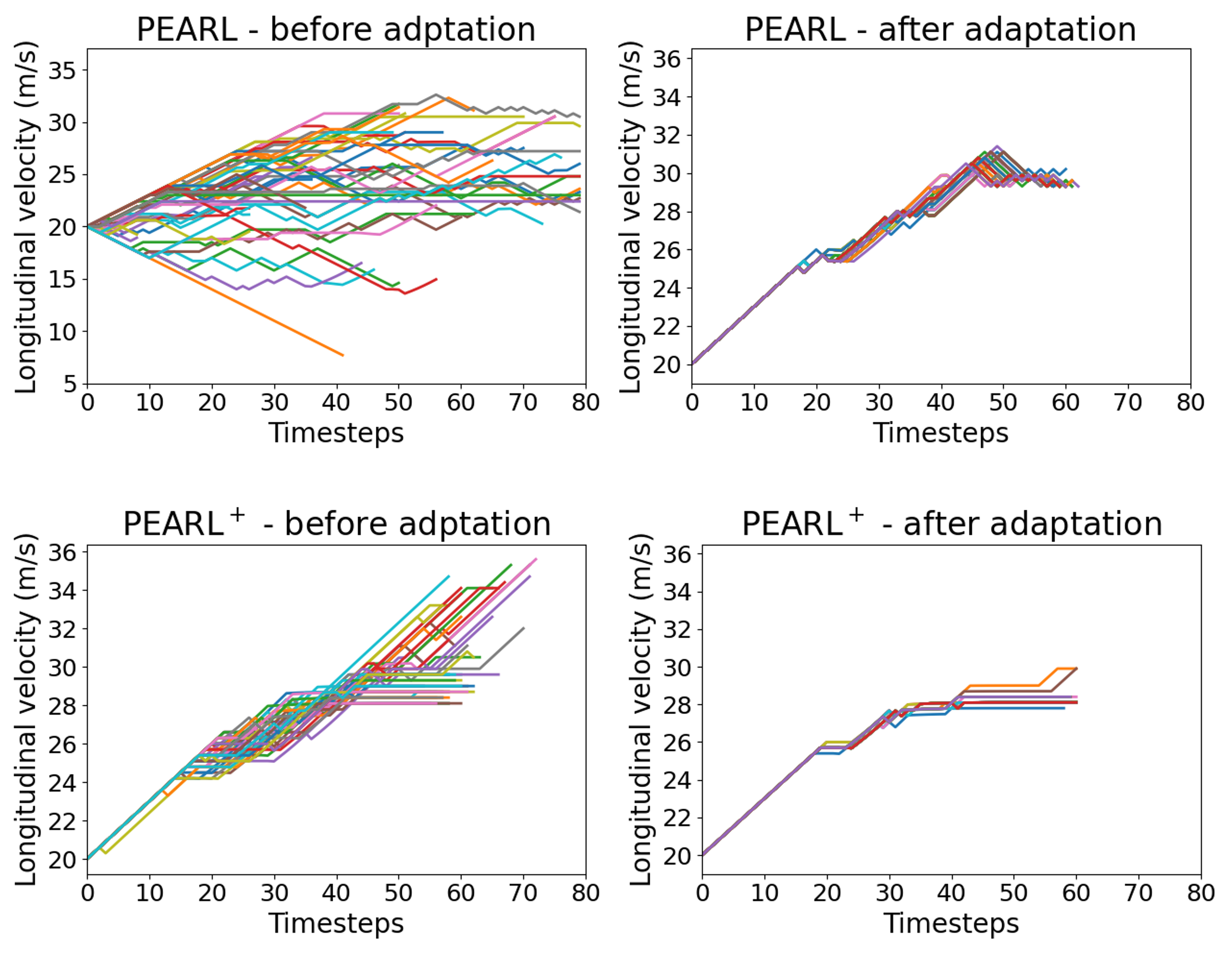}
	\caption{Exploration trajectories}
	\label{fig:traj}
\end{figure}

Each picture in Fig. \ref{fig:traj} plots 50 trajectories' speed profile for both before and after adaptation on a meta-testing task. The before-adaptation comparison between PEARL$^+$ and PEARL shows that PEARL$^+$'s exploration is more consistent and exhibits a safe and reasonable behavior (accelerating to catch up to the vehicles driving on the main-lane). After adaptation, both algorithms have similar performance.

\begin{figure}[ht]
	\centering
	\includegraphics[width=0.44\textwidth]{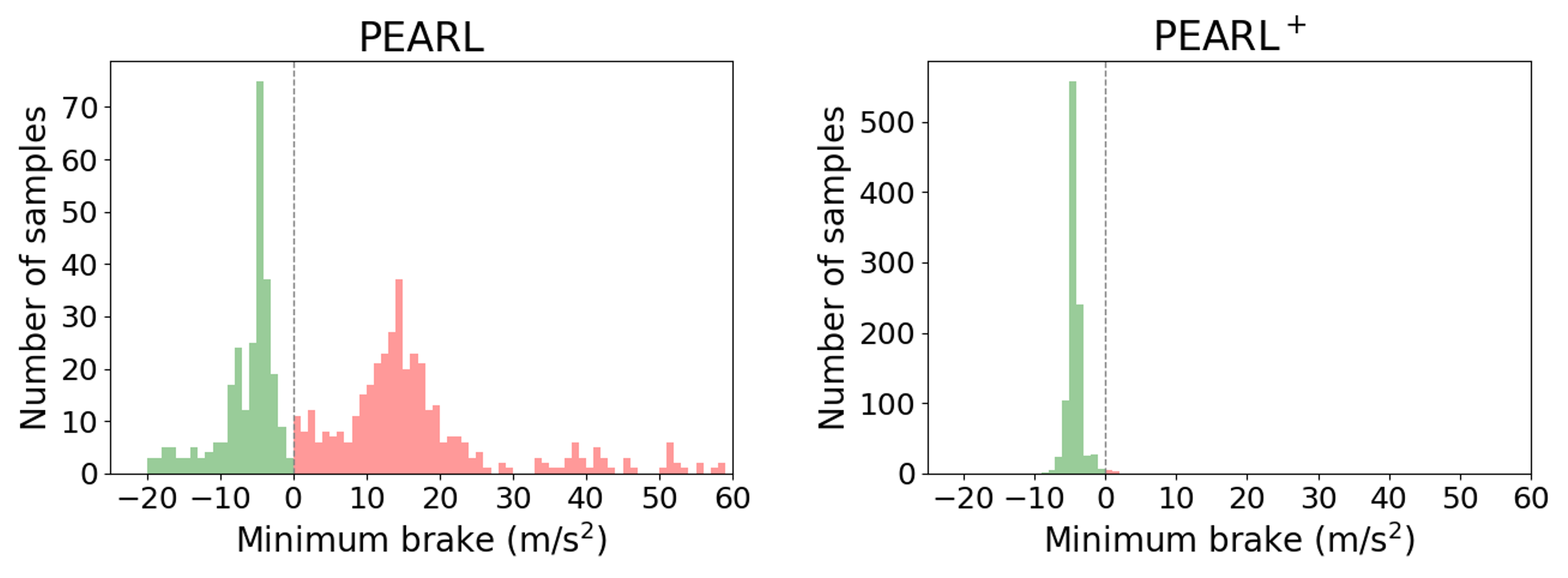}
	\caption{Minimum braking ($b_{min}$) required to avoid accident. (1000 roll-outs in total. \textcolor{green}{Green} represents negative $b_{min}$, \textcolor{red}{red} represents positive $b_{min}$), \textcolor{gray}{grey} dot-line is where $b_{min}=0$}
	\label{fig:min_brake}
\end{figure}

To compare the safety of the prior policy, we also analyze the minimum braking required to avoid crashes at the moment when the vehicle merges into the main lane in Fig. \ref{fig:min_brake}. The minimum braking $b_{min}$ is computed with the following equations:
\begin{equation}
\label{eq:min_b}
    b_{min} = \max{\{B_{min}(front, ego), B_{min}(ego, rear)\}}
\end{equation}
\begin{equation}
\label{eq:min_b_1}
    B_{min}(veh_0, veh_1) = \frac{v_1^2-v_0^2}{2(x_0-x_1)}
\end{equation}
where $x$ is the longitudinal position, $v$ is the velocity, and $front, rear, ego$ represents the front vehicle, rear vehicle and ego vehicle respectively.

The magnitude of positive value quantifies the braking effort required for the following vehicles (either ego or rear vehicles) and thus higher values imply higher likelihood of crashes (shown in red color in Fig. \ref{fig:min_brake}). In addition, negative $b_{min}$ means the following vehicle's speed is lower than the leading vehicle, thus is at low risk of crashes (shown in green color). We pay more attention to those cases with a positive $b_{min}$. From Fig. \ref{fig:min_brake} we can see that PEARL has more higher-risk cases compared with PEARL$^+$.

\begin{table*}[ht]
    \centering
	\caption{Meta-testing results and training efficiency comparison of the ablation study}
	\begin{subtable}{1.0\textwidth}
    	\centering
    	\begin{threeparttable}
    	\begin{tabular}{l|cc|cc|c}
            \hline
            \multirow{2}{*}{Ablation Algorithm} &
              \multicolumn{2}{c|}{Return} &
              \multicolumn{2}{c|}{Crash Rate} &
               \multirow{2}{*}{\tnote{1} Epoch \#} \\
                & \tnote{1}Before Adpt. & After Adpt. & Before Adpt. & After Adpt. &\\
                \hline
                PEARL & \underline{$45.48\pm 40.83$} & \underline{$106.36$}$\pm20.74$ & $33.33\pm 11.39\%$ & $\sim0.00\%$ & $\sim 275$\\
                \hline
                PEARL+ (w/o) & $92.47\pm 11.54$ & $151.32\pm 23.86$ & $\sim0.00\%$ & $ \sim0.00\%$ & $\sim 300$\\
                \hline
                PEARL+ (300-Node) & $\textbf{109.34}\pm 10.18$ &$\textbf{166.10}\pm 20.49$ & $\sim0.00\%$ &$\sim 0.00\%$  & $\sim 250$\\
                \hline
                PEARL+ (100-Node) & $\textbf{104.26}\pm 12.63$ &$\textbf{163.79}\pm 19.58$ & $\sim0.00\%$ &$\sim 0.00\%$  & $\sim \textbf{220}$\\
                \hline
    	\end{tabular}
    	\label{table:ablation_ant}
    	\caption{Ant3D-vel Experiment}
    	\end{threeparttable}
	\end{subtable}
	\begin{subtable}{1.0\textwidth}
    	\centering
    	\begin{threeparttable}
        	\begin{tabular}{l|cc|cc|c}
                \hline
                \multirow{2}{*}{Ablation Algorithm} &
                  \multicolumn{2}{c|}{Return} &
                  \multicolumn{2}{c|}{Crash Rate} &
                   \multirow{2}{*}{Epoch \#} \\
                    & Before Adpt. & After Adpt. & Before Adpt. & After Adpt. &\\
                    \hline
                    PEARL & \underline{$-33.44$}$\pm 3.41$ & $67.28\pm2.52$ & $27.84\pm 9.37\%$ & $\sim0.00\%$ & $\sim 400$\\
                    \hline
                    PEARL+ (w/o) & $42.47\pm $\underline{$18.13$} & $54.19\pm$\underline{$15.95$} & $0.74\pm0.95\%$ & $ \sim0.00\%$ & $\sim 300\pm 100$\\
                    \hline
                    PEARL+ (300-Node) & $\textbf{58.12}\pm 4.59$ &$\textbf{69.52}\pm 4.72$ & $0.86\pm0.75\%$ &$\sim 0.00\%$  & $\sim 430$\\
                    \hline
                    PEARL+ (100-Node) & $\textbf{56.47}\pm 3.88$ &$\textbf{70.14}\pm 3.26$ & $0.86\pm0.75\%$ &$\sim 0.00\%$  & $\sim \textbf{400}$\\
                    \hline
        	\end{tabular}
    	\label{table:ablation_highwaymerge}
    	\caption{Highway-Merge Experiment}
        	\begin{tablenotes}
        	    \item[1] The number of training epochs before weights of networks converge.
            	\item[*] Each mean and standard deviation is calculated on 4 independent runs.
            	\item[*] \textbf{Bold} means dominant advantages; \underline{Underline} means dominant disadvantages.
        	\end{tablenotes}
    	\end{threeparttable}
	\end{subtable}
	\label{table:ablation_exp}
\end{table*}

\subsection{Ablation study}
We conduct an ablation study to investigate the necessity of introducing an extra Q-network to estimate the action-state value with the prior context assumption. Because of large time and computation cost, we only carry out the ablation study on one Mujoco robotic task (Ant3D-vel) and the AV task (Highway-Merge). The four ablation algorithms we consider here are as follows:

\begin{itemize}
    \item PEARL: the original PEARL algorithm, which doesn't include a regularization term over the expected return of the policy conditioning on the prior context assumption.
    \item PEARL+ (w/o): this ablation algorithm introduces the regularization term concerning the policy with prior context assumption, but without introducing a new independent Q-network specially for state-action value estimation conditioning on prior assumption. In this setting, we calculate the loss $\Tilde{\mathcal{L}}_{critic}$ as $Q_{\theta}(\textbf{s, a, z}_0)$.
    \item PEARL+ (300-Node) and PEARL+ (100-Node): these two are both based on our proposed algorithm, including a regularization term in the objective function and a special Q-network to calculate the regularization term, but with different sizes of the extra Q-network. PEARL+ (300-Node) uses a 3-layer, 300-hidden-node Fully Connected Network (FCN) as the extra Q-Network, which is of the same size of Q-network used for the SAC policy training, while the 100-Node setting refers to using a 3-layer, 100-hidden-node FCN.
\end{itemize}

We use the same training hyper parameters for all these ablation algorithms, and list the meta-testing results and training efficiency comparison in Table \ref{table:ablation_exp}. In the following discussion, PEARL+ refers to both PEARL+(300-Node) and PEARL+(100-Node) as a whole.

Comparing the meta-testing results, we conclude that both PEARL+ and PEARL+(w/o) are safe before and after adaptation. However, PEARL+ gives better performance before adaptation and better adaptation ability. Specifically in the Highway-Merge experiment, we found that PEARL+(w/o) has higher variance than PEARL+, which is due to one time stuck at local optima. The superiority of PEARL+ compared to PEARL(w/o) might be because an extra Q-Network shares the burden of estimating state-action values, reducing the complexity of the action-state value function. Besides, the two-Q-Network PEARL+ framework can be viewed as division of the prior safety and adaptation ability: the original SAC Q-Network is responsible for adaptation (with the encoder), and the extra Q-Network is responsible for prior safety before adaptation. The division of Q-networks to consider the regularization term and the original optimization term might reduce the sensitivity to the weighting coefficient $\alpha$, whose effects will be discussed in the \ref{sec:alpha} section.

We further explore how the size of the extra Q-Network influence safety and adaptation ability. Comparing the meta-testing return of PEARL+(100-Node) and PEARL+(300-Node), we do not see obvious performance advantages, but PEARL+(100-Node) tends to require less epochs to converge than PEARL+(300-Node). This means that we can save training time and computational storage by using a smaller network to estimate prior state-action value.
\begin{figure}[h]
	\centering
	\includegraphics[width=0.28\textwidth]{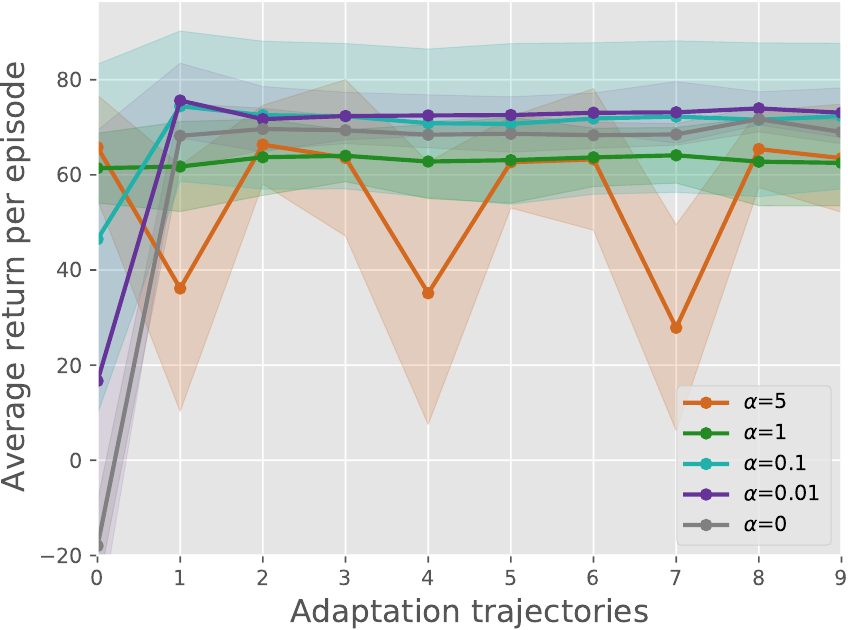}
	\caption{Meta-testing returns of Highway-Merge}
	\label{fig:alpha-meta-testing-returns}
\end{figure}
\begin{table}[H]
    \centering
	\caption{Meta-Testing crash rates of Highway-Merge}
	\label{table:Effect_of_additional_terms_weight}
	\begin{tabular}[ht]{M{0.1\linewidth}|M{0.25\linewidth}|M{0.25\linewidth}}
		\hline
		$\alpha$     & Before Adpt.         & After Adpt.   \\
		\hline
		5           &$\sim\textbf{0.00\%}$ &$11.8\%$ \\
        \hline
		1           &$0.5\%$        &$\mathbf{\sim 0.0\%}$ \\
		\hline
		0.1         &$2.5\%$        &$\mathbf{\sim 0.0\%}$ \\
		\hline
		0.01        &$35.0\%$       &$\mathbf{\sim 0.0\%}$ \\
		\hline
		0           &$37.5\%$       &$\mathbf{\sim 0.0\%}$ \\
		\hline

	\end{tabular}
\end{table}

\subsection{Effect of the weighting coefficient $\alpha$}
\label{sec:alpha}
In this section, we investigate the choice of the weighting coefficient of our additional optimization term in PEARL$^+$. Larger $\alpha$ results in more focus on the performance of the policy with prior, while smaller $\alpha$ optimizes more on the policy posterior.

Fig. \ref{fig:alpha-meta-testing-returns} indicates that an $\alpha$ that is too large can impair learning during adaptation (e.g., $\alpha=1$), and even fail to adapt ($\alpha=5$). On the other hand, an $\alpha$ that is too small will not achieve the desired prior policy safety, as shown in Table \ref{table:Effect_of_additional_terms_weight} ($\alpha=0.01$). Therefore, the weight needs to be chosen to balance between prior safety and posterior adaptation. In our case, $\alpha=0.1$ is a good choice in all three experiments, but it is possible a different $\alpha$ needs to be used for other applications.

\section{Conclusion}
\label{Conclusion}
In this paper, we propose a Meta-RL algorithm that considers pre-adaptive safety and robust performance to unseen tasks. This is an essential feature for safety-critical real-world applications. By introducing an additional regularization term over the expected return of the policy with prior context assumption, and a corresponding additional critic network that returns the Q value conditioned on the prior context, our algorithm PEARL$^+$ achieves superior performance (mostly through better safety reward) when it is exposed to a new task for the first time.  In addition, it preserves the fast adaptation of PEARL, the off-policy Meta-RL method our algorithm builds upon. Our algorithm has been validated using three example applications: two MuJoCo robotic tasks and a highway merge autonomous driving task. 
These experiments show that PEARL$^+$ achieves better robustness and safety performance, and the ablation study demonstrates the necessity of introducing an extra Q-Network. Furthermore, the prior regularization term is not restricted to context-based Meta-RL algorithms, and actually can be applied to any other Meta-RL frameworks, which would be a good direction for future work 

\bibliography{main.bib}{}
\bibliographystyle{ieeetr}

\end{document}